\documentclass[10pt,twocolumn,letterpaper]{article}

\usepackage[pagenumbers]{cvpr} 

\usepackage{graphicx}
\usepackage{amsmath}
\usepackage{amssymb}
\usepackage{booktabs}

\usepackage[pagebackref,breaklinks,colorlinks]{hyperref}

\usepackage{tabularx}

\usepackage{enumitem}
\usepackage{array}
\usepackage{balance}
\usepackage{color}
\usepackage{xcolor}         
\usepackage{bbm}
\usepackage{algorithm}
\usepackage[noend]{algpseudocode} 
\usepackage{bbding}
\usepackage{booktabs}
\usepackage[normalem]{ulem}
\usepackage{setspace}
\usepackage{stfloats}
\usepackage{bm}

\usepackage{amsmath}
\usepackage{amssymb}
\usepackage{amsfonts}       
\usepackage{comment}
\usepackage{nicefrac}       
\graphicspath{ {./Figures/} }
\usepackage{lipsum}
\usepackage{etoolbox}
\usepackage{geometry}
\usepackage{indentfirst} 
\usepackage{ulem}
\usepackage{graphicx}
\usepackage{soul}
\usepackage{dsfont}

\let\tinymatrix\smallmatrix

\patchcmd{\tinymatrix}{\scriptstyle}{\scriptscriptstyle}{}{}
\patchcmd{\tinymatrix}{\scriptstyle}{\scriptscriptstyle}{}{}
\patchcmd{\tinymatrix}{\vcenter}{\vtop}{}{}
\patchcmd{\tinymatrix}{\bgroup}{\bgroup\scriptsize}{}{}

\usepackage[overload]{textcase}

\newcolumntype{C}[1]{>{\centering\let\newline\\\arraybackslash\hspace{0pt}}m{#1}}

\def\cvprPaperID{57} 
\def\confName{CVPR}
\def\confYear{2022}

\begin{document}

\title{Spatial Cross-Attention Improves \\ Self-Supervised Visual Representation Learning}

\author{Mehdi Seyfi, Amin Banitalebi-Dehkordi, and Yong Zhang\\
Huawei Technologies Canada Co., Ltd.\\
}

\maketitle

\begin{spacing}{1.00}

\begin{abstract}
Unsupervised representation learning methods like SwAV \cite{caron2020unsupervised} are proved to be effective in learning visual semantics of a target dataset.
The main idea behind these methods is that different views of a same image represent the same semantics.
In this paper, we further introduce an add-on module to facilitate the injection of the knowledge accounting for spatial cross correlations among the samples. This in turn results in distilling intra-class information including feature level locations and cross similarities between same-class instances. The proposed add-on can be added to existing methods such as the SwAV.
We can later remove the add-on module for inference without any modification of the learned weights. Through an extensive set of empirical evaluations, we verify that our method yields an improved performance in detecting the class activation maps, top-1 classification accuracy, and down-stream tasks such as object detection, with different configuration settings. 
\end{abstract}
\section{Introduction}


Self supervised representation learning has recently gained a dramatic attention as a solution to combat the hassle of collecting and annotating large datasets. 
In contrast to semi-supervised learning, self supervised methods do not utilize any labels for training models \cite{xmoco_ieee,akbari2022lang,banitalebi2021repaint,banitalebi2021knowledge,banitalebimodel,banitalebi2021auto,ramamonjison2021simrod}.
The fundamental approach in self supervised learning is to design a so-called \textit{pretext} task that does not require human-labelled annotations, but instead can generate the necessary supervisory information from the training data itself. Initial attempts in this regard included pretext tasks such as rotation classification~\cite{rotation}, solving a jigsaw puzzle~\cite{noroozi2016unsupervised}, image colorization~\cite{zhang2016colorful}, inpainting~\cite{pathak2016context}, etc. 


A breakthrough in self supervised learning however, occurred by the emergence of contrastive learning~\cite{Wu_2018_CVPR} and infoNCE loss~\cite{oord2018representation,henaff2019data}. The contrastive representation learning idea is to train a network that is able to discriminate between the images in a dataset, via minimizing the similarity between their representation features through the infoNCE loss. Later, \cite{Misra_2020_CVPR} and \cite{cvpr19unsupervised} reasoned that useful image representations must be invariant under semantics-preserving image transformations. This observation formed the foundation of many newer State-Of-The-Art (SOTA) algorithms in the filed. To comply with this observation these algorithms force the representations of the transformed image counterparts to be close in terms of their normalized inner product, while at the same time discriminate them against negative samples within the dataset by minimizing their cosine similarity~\cite{he2019moco, chen2020mocov2,chen2020big,chen2020simple}.
Later on, Siamese networks started another trend in representation learning by eliminating the need for negative samples~\cite{caron2020unsupervised,grill2020bootstrap,chen2020exploring}. 


The foundation of both contrastive and Siamese based methods is set upon the transformation invariance argument~\cite{Misra_2020_CVPR,cvpr19unsupervised}. These methods learn from enforcing dissimilar views to be similar in the embedding domain, and thus do not consider the locality information in the transformed image pairs~\cite{he2019moco, chen2020mocov2,chen2020big,chen2020simple,caron2020unsupervised,grill2020bootstrap,chen2020exploring,asano2019self}. In other words, they ignore the spatial location of the target entity/object in a scene. The other drawback is the random/multi crop augmentation used in their image transformation pipelines. Random multi crop transformations are used based on the assumption that different crops of a same image represent the same high-level semantics, which may not necessarily be true, especially for cases where the object of interest does not appear (even partly) in the generated crops. These issues are exacerbated in complex scenes with multiple objects, or when the foreground object comprises a small portion of the scenes. Therefore, occurrence of these circumstances question the efficacy of the existing methods for complex scenes or non-classification down-stream tasks \cite{henaff2021efficient, xie2021detco, yang2021instance}. 

\begin{figure*}
     \centering
     \subfloat[The scene contains multiple object, in this scenario random crops of the same image may carry different context.]{
         \centering
        \includegraphics[width=12.5cm]{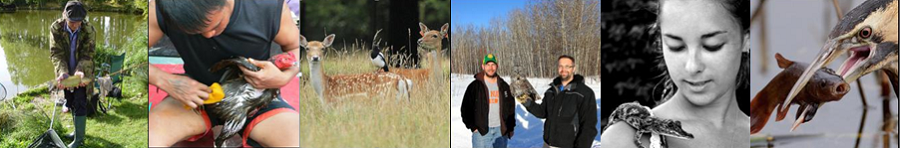}
         \label{fig:multi_obj}}
     \hfill
     \subfloat[The background is dominant in the scene. Random crops may not contain the foreground object.]{
         \centering
         \includegraphics[width=12.5cm]{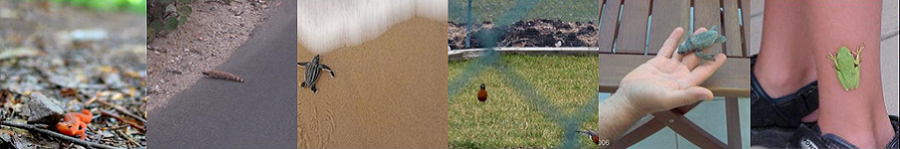}
         \label{fig:small_obj}}
         \vspace{-8pt}
\caption{Examples where random crop in self representation learning methods may fail. Images from Imagenet-1K.}
\end{figure*}





In this paper, we argue that in the self representation learning methods, it is important for the model to know where to focus for the foreground object in each training instance. Key information elements are the features associated to the foreground object in the feature space, and key regions are the areas corresponding to the key features.
Key features carry all the information about the context of the instance.
We emboss these features by cross correlating the visual patterns of instances in the dataset that are estimated to be in the same category (positive images)\footnote{Positive images are not necessarily different instances. They can be different crops of the same image.}. In other words, by searching for shared visual patterns across the same-class instances, we localize and highlight the spatial extent of the foreground objects within positive images. After learning the where-to-look masks from cross correlating the visual patterns associated to the positive instances, we generate explanation maps thereby embossing the commonalities and muting the other parts across the positive feature maps. We then increase the output confidence by traversing the feature maps super-imposed with the highlighted key regions, i.e., explanation maps, through the network~\cite{chattopadhay2018grad,wang2020score}. We finally match the output scores with pseudo labels achieved from clustering the representation features.


We introduce an add-on to the existing Siamese/contrastive~\cite{chen2020exploring,caron2020unsupervised,chen2020mocov2,he2019moco} architectures that can be trained in a completely self-supervised manner. This add-on can later be detached from the backbone in the inference time or when fine-tuning the backbone.
By incorporating this attention-like module, the network extracts additional knowledge about the true semantics of the images and therefore premises its decision logic upon salient features of the incoming data.


Our results show that models trained with our method can generally exhibit a better interpretation. The aim of model interpretation is to describe the logic behind why a model decides a particular decision in a downstream task. Via studying model interpretation we argue that self supervised deep
networks may make decisions based on irrelevant representations. We study the model interpretation via looking at the heatmaps of the finetuned classification model. The heatmaps reveal the areas that the model pays attention to, when deciding an outcome at the inference. We also quantitatively study the performance of our proposed method in incorporating key features in the model interpretation by measuring the Average Drop (AD) and Average Increase (AI) values\cite{chattopadhay2018grad}. 

The main contributions of this paper can be summarized as follows:
\begin{itemize}
\item We introduce an add-on for the current self supervised representation learning methods that can be detached from the baseline model in the inference or downstream task fine-tuning.
\item The proposed add-on injects extra information about the location of the foreground objects in the training instances, via which the model reveals a better decision making mechanism. We verify our claims with the widely accepted interpretation metrics, AD, and AI.
\end{itemize}

\section{Method}\label{sec_method}
\begin{figure*}[h]
    \centering
    \includegraphics[width=\textwidth]{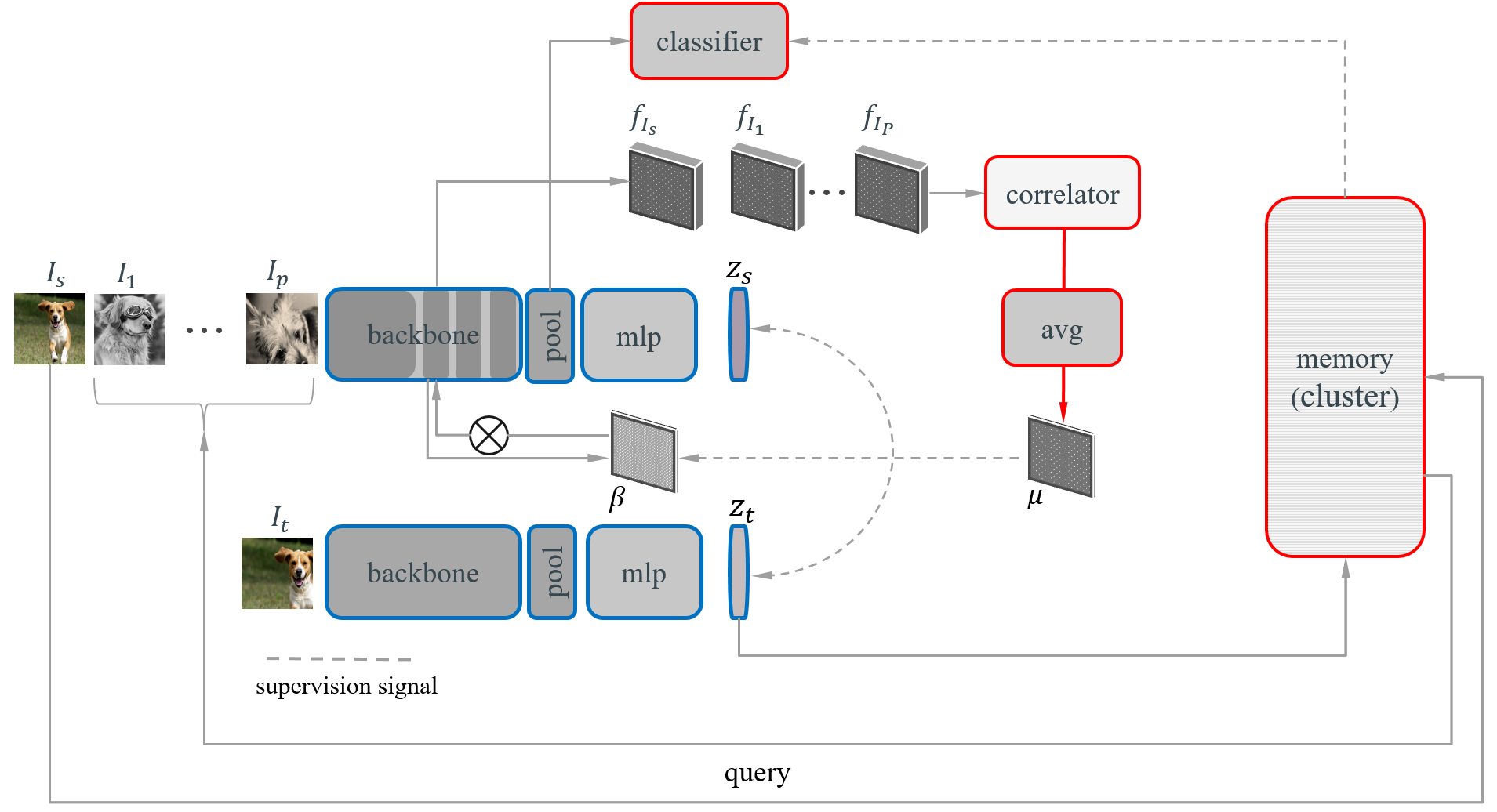}
    \caption{Schematic of the proposed method. The core engine is outlined by blue and the attention module with red. Attention mask $\beta$ predicts the correlation mask $\mu$ via a simple convolutional layer. The correlation between the feature maps $f$ of the images in the the set $\{I_s, \mathcal{P}\}$ generates a correlation mask $\mu$ that highlights key regions in the images from the same cluster. Then $\beta$ is multiplied to the feature maps to generate explanation maps. The explanation maps are then passed through a pooling layer and a fully connected layer to predict the labels of the images in $\{I_s, \mathcal{P}\}$. The network is trained with the pseudo-labels obtained by clustering the memory bank.}
    \label{fig:total}
\end{figure*}

In this section, we first describe a formulation of the problem along with some basic setup, and then explain our method and its components.

A typical SSL system consists of a backbone network $\mathcal{B}_{\circ}$,  a pooling layer $\mathcal{P}_{\circ}$, a MLP head $\mathcal{M}_{\circ}$, an encoder network $\theta$, and a classification head $\mathcal{C}_{\circ}$. 
Let us denote the input image dataset by $\mathcal{I}$ and a set of random transformation functions by $\mathcal{T}_{\circ}(\cdot)$. We define the backbone network by  $\mathcal{B}_{\circ}:\mathcal{I}\longrightarrow f \in \mathds{R}^{D\times H\times W}$, where $D$, $H$, and $W$ denote channel depth, height, and width of the backbone output feature map $f$, respectively. The features are then forwarded to
a global average pooling layer $\mathcal{P}_{\circ}:f\longrightarrow \mathds{R}^D$, followed by a MLP head $\mathcal{M}_{\circ}:\mathds{R}^{D\times H\times W}\longrightarrow z \in \mathds{R}^d$, where $d$ is the embedding dimension. Suppose the final classification head is denoted by $\mathcal{C}_{\circ}:\mathds{R}^d\longrightarrow \mathds{R}^C$, where $C$ is the number of classes. Then, the encoder network defined by $\theta(\cdot) \triangleq \mathcal{M}_{\circ}(\mathcal{P}_{\circ}(\mathcal{B}_{\circ}(\cdot)))$ generates representation vectors $z$ from the input image $I$. 

In our method, we define an add-on attention module $\mathcal{A}_{\circ}:f\longrightarrow \beta \in \mathds{R}^{1\times H \times W}$, and apply its output as mask weights to the feature maps. In addition, we cluster the representation vectors and use the clusters as pseudo labels to minimize the following loss function:
{
\begin{eqnarray}
\mathcal{L}_{\texttt{cls}} = &-&\mathds{E}_{\begin{tinymatrix}\!\!\!\! I\in \mathcal{I}\\t\in\mathcal{T}_{\circ} \end{tinymatrix}}\bigg\{
\sum_{y=1}^{C}\delta(y-\hat{y}_{I})\\ &&\cdot\log\Big(\mathcal{C}_{\circ}\left(\mathcal{P}_{\circ}\left(f\odot\beta\right)\right)[y]/\tau_c\Big)\bigg\},\nonumber
    \label{eq_system_model}
\end{eqnarray}
}
where $\hat{y}_I$ is the pseudo label generated by clustering the representation vectors $z$ associated with image $I$, $\tau_c$ is a temperature value, and $\odot$ denotes the element-wise multiplication. In the following subsections, we describe different components in more details, and explain how we solve (\ref{eq_system_model}).
\subsection{Clustering}
At the beginning of each epoch, we first compute the representation vectors $z=\theta(\mathcal{T}_{\circ}(I)) \in \mathds{R}^{d}$ for $\forall I\in {\mathcal{I}}$, and store them in a memory bank which is registered to contain the representation vectors of the dataset $\mathcal{I}$.   
The memory bank is then clustered to $C$ classes of representations using the K-means algorithm \cite{caron2018deep}. This way each $I \in \mathcal{I}$ is associated with a pseudo label $\hat{y}_I$ generated via clustering.
\subsection{Canonical representation}
In the second step, a source image $I_s=\mathcal{T}_{\circ}(I)$ and its transformed (augmented) view $I_t=\mathcal{T}_{\circ}(I)$, are fed to a canonical representation learning method with transform consistency loss i.e., the output representation of transformed views of an image are enforced to converge to each other via some similarity function $\mathcal{S}$,
\begin{eqnarray}
\mathcal{L}_{\texttt{ssl}} &=& \frac1{2}\mathcal{S}\big(h\big(\theta(I_s)\big), \theta(I_t)\big) \nonumber\\&+& \frac1{2}\mathcal{S}\big(h\big(\theta(I_t)\big), \theta(I_s)\big),
\label{eq_ssl}
\end{eqnarray}
where $h$ can be an MLP prediction head or an identity network depending on the self supervised core-engine \cite{he2019moco,chen2020simple,caron2020unsupervised,grill2020bootstrap}.
The canonical representation learning method could come from an existing approach such as Siamese \cite{caron2020unsupervised, chen2020exploring} or contrastive \cite{he2019moco,chen2020simple} techniques. In our experiments, we implement our method on top of SwAV \cite{caron2020unsupervised} and SimSiam \cite{chen2020exploring} core representation learning engines however, results can be extended to other contrastive methods such as \cite{he2019moco, chen2020mocov2,chen2020big} as well.
\subsection{Attention module and positive sampling}
By generating attention-like masks, we incorporate the location information of the regions in image $I\in\mathcal{I}$ as a key factor that can enhance the mutual information between $I$ and and its representation vector $z=\theta(\mathcal{T}_{\circ}(I))$ \cite{noroozi2016unsupervised}. On the other hand, as reported by \cite{hock1974contextual}, contextual information between the objects of same category can influence the performance of object recognition. Therefore, in our method we propose a positive image sampling procedure via which we search over commonalities among the the positive image set in the feature space. Then we estimate an attention mask thereby highlighting the most correlated regions among the positive set as the regions of interest. Intuitively, we argue that the common visual similarity between a set of positive images, contains key information about the context of objects in the scene. Through this mechanism we redirect the attention of the network to common features in the positive set on one-hand, and to the location of these features, on the other hand. This is done by multiplying the normalized attention mask to the feature maps generated by the network. In the following, we will explain in details the process of learning the attention-like masks.

At first, each image $I_s$ in a batch is assigned a pseudo-label (retrieved by clustering the memory bank at the beginning of each epoch). We then run an enquiry for each $I_s$, and sample $P$ similar images based on their clustered pseudo-labels in the memory. The positive set $\mathcal{P} = \left\{ \mathcal{T}_{\circ}(I_1),\mathcal{T}_{\circ}(I_2) \ldots,\mathcal{T}_{\circ}(I_P)\right\}$ is constituted from images in the same cluster as $I_s$.  

Next, for each image $I$ in the image set $\left\{I_s, \mathcal{P}\right\}$, we get the feature map $f_I=\mathcal{B_\circ}(I)$ from an intermediate bottleneck layer of the backbone network, with $[f_I]_{ij}\in \mathds{R}^{1\times D}$ representing the context feature vector at spatial location $(i,j)$. These feature vectors encode spatio-contextual information about the input image $I$. 

Looking for interconnections between the input image $I_s$ and the positive set, we measure the contextual overlaps between $f_{I_s}$ and the positive set feature maps by correlating the feature vectors in $f_{I_s}$ and the positive set members (denoted by $I^{'}$) \cite{zhou2020look}, as follows:
\begin{equation}
\rho_{ij}(I_s, I^{'}) = \frac{1}{C} \max_{\begin{tinymatrix}0\leq i^{'}<H\\0\leq j^{'}<W\end{tinymatrix}}
 [f_{I_s}]_{_{ij}} [f_{I^{'}}]_{_{i^{'}j^{'}}}^{T}.\label{eq_corr}
\end{equation}
By concatenating the correlation scores $\rho_{ij}(I_s, I^{'})$ in a $H\times W$ grid, we generate a correlation mask which supposedly encodes the intersectional contexts between $I_s$ and $I^{'}\in \mathcal{P}$. Note that the $\max$ operator in \eqref{eq_corr} embosses the most attentive common semantics between $I_s$ and $I^{'}$. We resume by generating the correlation mask $\rho$ for each $(I_s, I^{'})$ pair in the positive set $\mathcal{P}$. Finally, we average the correlation masks across the positive set to obtain a mask corresponding to $I_s$:
\begin{equation}
    \mu(I_s, \mathcal{P}) = \frac1{P}\sum_{p=1}^P \rho(I_s, I^{'}_p).\label{eq_corr_mask}
\end{equation}
Since images in $\left\{I_s, \mathcal{P}\right\}$ are subject to transformations such as $\texttt{RandomResizedCrop}$, we expect the correlation mask $\mu(I_s, \mathcal{P})$, to capture the locality as well as the common semantics between the images in the same cluster. We therefore, reinforce the model weights by fusing the semantic information acquired by the correlation mask. To do so, we detach the gradient flow from the correlation mask and use it as a supervision signal to train the backbone feature maps. To match the shapes and dimensions, we attach a $1\times 1$ convolutional layer preceded by a $\texttt{Relu()}$ activation to the backbone. Moreover, we use a mean square error (MSE) function to minimize:
\begin{equation}
\mathcal{L}_{\mu} = \frac1{H.W}\sum_{i}\sum_{j} \left[\beta_{ij} - \mu_{ij}(I_s, \mathcal{P})\right]^2,     \label{eq_mse}
\end{equation}
where $\beta$ is the attention mask obtained by passing the feature map $f$ through a $\texttt{Relu()}$ layer followed by a $1$ strided convolutional layer with $D$ input channels and a single output channel.
The effect of the correlation mask $\mu(I_s, \mathcal{P})$ is slid through the network by back-propagating the gradient of the loss function \eqref{eq_mse} through the backbone. 

Additionally, we generate explanation maps by multiplying the captured attention mask $\beta$, to the feature maps $f_I$, for $I\in\left\{I_s, \mathcal{P}\right\}$. The explanation maps then traverse through the rest of the network, pass a pooling layer and finally a fully connected layer with $C$ output neurons. The effect of the attention map on the feature maps is distilled into the explanation maps and subsequently the whole network through a weakly-supervised cross entropy loss like \eqref{eq_system_model} as:
\begin{equation}
    \mathcal{L}_{\texttt{cls}} = -\frac1{N}\sum_{n=1}^N\sum_{y=1}^{C}\delta(y-\hat{y}_{I_n}) \log(\mathbf{c}_n[y]/\tau_c),\label{eq_cls_loss}
\end{equation}
where $N$ is the batch size, $\mathbf{c}_n$ is the classification head output logit vector associated to sample $n$ in the mini batch, $\tau_c$ is the temperature value that controls the output smoothness, and $\hat{y}_{{I}_n}$ corresponds to the pseudo-label of $I_n\in\mathcal{I}$ acquired by the k-means clustering of the memory bank. 
The attention map $\beta$, highlights the commonalities between images in the same cluster. The explanation map extracted this way, helps the network to increase the accuracy of classification as only the important foreground information is fed to the the classifier. On the other hand, regions with less common semantics with the key image $I_s$ are automatically muted. Therefore, if an image crop does not contain the foreground regions, then its corresponding attention mask is consistently muted, which results in a lower impact in the classification loss of \eqref{eq_cls_loss}. This means that, the attention module does not affect the network weights with the image crops that do not contain foreground semantics.
Figure \ref{fig:total} shows how the proposed add-on is appended to the core representation learning engine.
Putting all together, the overall loss can be expressed as:
\begin{equation}
    \mathcal{L} = w_0 \mathcal{L}_{\texttt{ssl}} + w_1 \mathcal{L}_\mu + w_2 \mathcal{L}_{\texttt{cls}}, \label{eq_final_loss}
\end{equation}
where $w_i$s are weights assigned to different terms.

\section{Experiments details}
Previous self-supervised methods report unsupervised learning accuracies with different structures, number of epochs trained and architecture sizes. In this paper we choose the common Resnet-50 structure and compare our results with the baseline architecture in for 200 epochs of pre-training on Imagenet-50 and Imagenet-1K. 

More information about the details of training can be found in the supplementary material.

\subsection{Classification, K-Nearest Neighbors }\label{sec_knn_classification}
After each epoch of training, we monitor the performance of the model weights on the test dataset, using a k-nearest neighbors (KNN) monitor\cite{Wu_2018_CVPR} with $k=200$. KNN accuracy could be used as a metric of how well the representation vectors of same classes can be grouped together. For this part after each epoch we store the representation vectors of each data point in the training set in memory and will match the test set data points to the stored representation vectors via a weighted KNN algorithm. Table.~ \ref{tab:knn_Imagenet} compares KNN accuracies of the baseline SwAV and our method, for Imagenet-1K and Imagenet-50 datasets.
\begingroup
\setlength{\tabcolsep}{1pt} 
\renewcommand{\arraystretch}{1} 

\begin{table}[htb!]
    \centering
\begin{tabular}{c|c|c|c|c}
 Algorithms & acc  & epochs & batch size& dataset\\
\hline
SwAV+repo & 46.5   & 200 & 4096&Imagenet-1K\\
\hline
Ours & $\mathbf{50.2}$  &200 & 2048&Imagenet-1K\\
\hline
SwAV+repo & 52   & 200 & 256&Imagenet-50\\
\hline
Ours & $\mathbf{65}$  &200 & 256&Imagenet-50\\
\hline
\end{tabular}
\caption{KNN accuracy with 200 neighbors on Imagenet-1K and Imagenet-50 with Resnet-50 backbone.}
    \label{tab:knn_Imagenet}
\end{table}
\endgroup
\subsection{Classification, the linear head protocol }\label{sec_linear_classification}
After pre-training the network unsupervisedly we train a linear head on top of the backbone as a conventional practice in self-supervised learning on both Imagenet-50 and Imagenet-1K datasets. For details please refer to supplementary material. 


Table.~\ref{tab:linear_Imagenet} shows the top-1 accuracy comparison of SwAV and our proposed model. Our results show improvements on both Imagenet-1K and Imagenet-50 datasets; however the improvement in the Imagenet-50 is more dramatic.
\begingroup
\setlength{\tabcolsep}{1pt} 
\renewcommand{\arraystretch}{1} 
\begin{table}[htb!]
\centering
\begin{tabular}{c|c|c|c|c}
 Algorithms & acc  & epochs & batch size &dataset\\
\hline
SwAV+repo & 68.4  & 200 & 4096& Imagenet-1K\\
\hline
Ours & $\mathbf{68.9}$ & 200 & 2048&Imagenet-1K\\
\hline
SwAV+repo & 68.2  & 200 & 4096&Imagenet-50\\
\hline
Ours & $\mathbf{75}$ & 200 & 2048&Imagenet-50\\
\hline
\end{tabular}
\caption{Top-1 accuracy for the conventional linear head protocol when a Resnet-50 backbone trained on Imagenet-1K and Imagenet-50 datasets.}
\label{tab:linear_Imagenet}
\end{table}
\endgroup
\subsection{Classification, Pascal-VOC}
\begin{table*}[htb!]
    \centering
\begin{tabular}{c|c|c|c|c}
Algorithms  & \texttt{val2012} AP & \texttt{test2012} AP & arch & epochs\\
\hline
Imagenet-1K-pretraining  & 92.0& 93.2 & Res-50 &  \\
\hline
SwAV-Pretraining on Imagenet-1K & 90.5& 91.0 & Res-50 &200 \\
\hline
Ours-Pretraining on Imagenet-1K & 92.4& 93.1 & Res-50 & 200\\
\hline
SwAV trained on Pascal & 71.7& 73.1 & Res-50 & 200\\
\hline

Ours trained on Pascal & 73.5& 74.5 & Res-50 & 200\\
\hline
\end{tabular}
\caption{Comparison between Imagenet-1K pretraining vs SSL training from scratch on Pascal-VOC. As it can be seen, the performance drops significantly when trained on Pascal-VOC unsupervisedly. This corroborates that augmentation-invariance fails to learn good representations via random cropping of image views in SSL.}
    \label{tab:pascal_ssl_cls_10_epoch}
\end{table*}

In this section we train an end-to-end Resnet-50 model on Pascal-VOC \cite{Everingham10} dataset for 10 epochs. A linear head to classify 20 objects of Pascal-VOC dataset, is appended to a pretrained backbone. 
We use batch size 16 with $lr= .00015$ for the backbone and $lr=.05$ for the linear head. We fine-tune the pretrained model only for 10 epochs to make sure that the initial backbone weights do not change drastically. For details of the training procedure please refer to the supplementary material.
The goal of this task is to predict the presence/absence of at least one object in each test image. Therefore, training 20 separate object classifiers, we use \texttt{BCEWithLogitsLoss} as our loss function and report the average precision as a the performance metric.
For this experiment we train the backbone in two different scenarios: \\
\textbf{Self supervised pretraining on Imagenet-1K:} The objective of this scenario is to verify the transferability of the pretrained model wieghts from Imagenet-1K to Pascal-VOC, when the backbone is pretrained in a self supervised manner. Table.~\ref{tab:pascal_ssl_cls_10_epoch} shows that self supervised pretraining performs competitively with its Imagenet-1K pretrained counterpart. More importantly, when equipped with our proposed add-on, we outperform SwAV on both validation and test sets for $\sim+2\%$ in terms of average precision.\\  
\textbf{Self supervised pretraining on PASCAL-VOC}: This scenario experiments the effectiveness of self supervised pretraining on complex datasets like Pascal-VOC. As table.~\ref{tab:pascal_ssl_cls_10_epoch} shows, there is a $20\%$ drop in the average precision of Pascal-VOC classification, if the backbone is pretrained in a self supervised manner on the very same dataset instead of Imagenet-1K. One reason is the amount of Imagenet-1K data. Imagenet-1K has $\times 10$ more data points as much as Pascal-VOC, this indeed contributes to the fine-tuned average precision. The other reason though is the complexity of Pascal-VOC dataset. In each training image, there are multiple objects; therefore, augmentation invariance principle deteriorates the ability of the model to learn proper representations via maximizing the cosine similarity of the representation vectors associated to random crops of the same image. The reason is random crops of the same image may contained different objects of the same image. Therefore, as we argued before, injecting information about the contextual extent of each object in the image will assist the core engine to learn better representations and transfer better to a multi-object classification downstream task. Table.~\ref{tab:pascal_ssl_cls_10_epoch} shows that our add-on improves SwAV's performance for $+1.8\%$ on the validation set and $+1.3\%$ on the test set in terms of average precision.

\subsection{Interpretability}
To understand the internal decision making process and the logic behind the network inference, one solution is to utilize class activation map (CAM) \cite{zhou2016learning,wang2020score}. CAM is a technique that provides a visual explanation of how a weighted linear combination of convolutional layer activation maps can render valuable information about specific region of the input data that the model mostly relies on in its decision making process. In a simple word CAM indicates the region that the network is paying attention to, when making a decision.

There are various version of CAM technique available in the literature \cite{zhou2016learning,wang2020score,chattopadhay2018grad,selvaraju2017grad}, however assuming a model with a trained linear head, a simple proposition of CAM for a class of interest $c$ can be defined as \cite{wang2020score}
\begin{equation}
    L_{\texttt{CAM}}^c = \textrm{ReLU}\left(\sum_{j} \alpha_j^c A_j\right),\label{eq_cam}
\end{equation}
where $A_{j}$ is the $jth$ channel activation in the last layer of the backbone network before the pooling layer, and $\alpha_j^c$ is the element at the $cth$ row and $jth$ column of the  weight matrix connecting the pooling layer to the output fully connected layer. 
\begin{figure*}[htb!]
\centering
\includegraphics[width=1\textwidth]{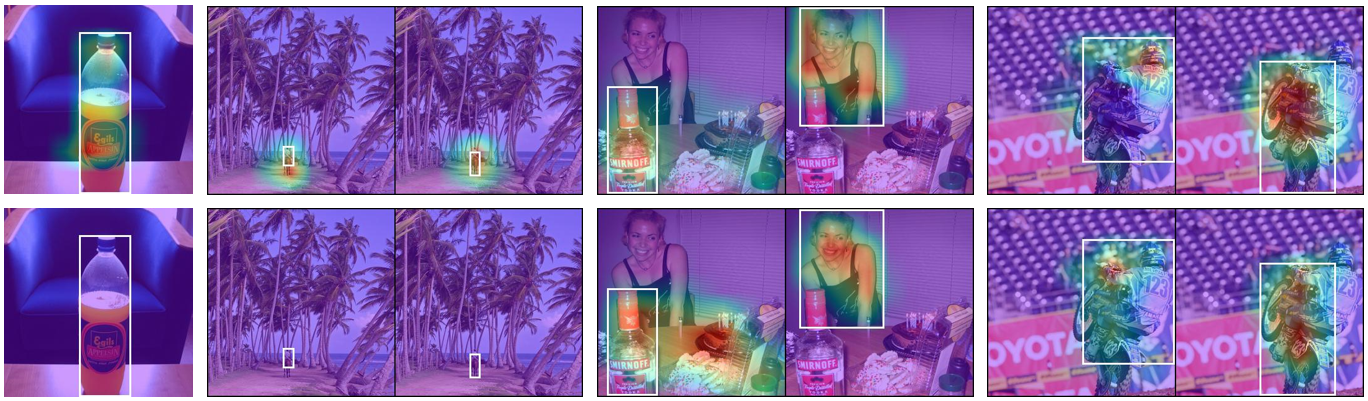}
\caption{Our method (top row) does a better job on focusing the more salient pixels than SwAV (bottom row). The ground truth bounding boxes are shown on each image. The Images are taken from Pascal-VOC dataset 2007.}
\label{fig:saliency_map}
\end{figure*}
We run a set of experiments to show that our methodology does a better job in interpreting the input data.
Using CAM, we evaluate these interpretations via extracting class specific saliency maps generated by the model when exposed to the input image data. Explanation maps are generated thereby multiplying the resized then normalized saliency maps to the input images.

Via a set of rules, the network makes a particular decision on a downstream task based on salient regions in the input image. The notion of salient regions can be quantitatively expressed as an average drop in the output confidence score if these regions are completely or partly muted. Average drop, AD,  is defined as \cite{chattopadhay2018grad}
$$
AD = \sum_{n=1}^N\frac{\texttt{max}\left(0, Y_i^c- O_i^c\right)}{Y_I^c}\times 10,.
$$       
where $Y_i^c$ is the prediction score of class $c$ for image $i$ in the dataset, and $O_i^c$ is the prediction score when the explanation map is fed to the model. On the contrary, if the salient region of the input data is embossed relative to other non-salient regions an increase in the confidence score is expected. This is expressed via defining average increase, AI, \cite{chattopadhay2018grad}
$$
AI = \sum_{i=1}^N\frac{\texttt{sgn}\left(Y_i^c - O_i^c\right)}{N},
$$
where $\texttt{sgn}$ is the sign function.

We run a set of experiments on the full Imagenet-1K/Imagenet-50 validation datasets. After a linear head is fine-tuned on Imagent-1K, we extract the saliency maps associated to the ground truth labels $c$ via \eqref{eq_cam}. We compare the output confidence score of the input images vversus their associated explanation maps via AD and AI metrics. Table.~\ref{tab:quantitative} shows $\simeq+4\%$ improvement on the average increase.

Additionally Fig.~\ref{fig:saliency_map} shows a comparison with the saliency maps between our method and that of SwAV's. As it is shown in the sample images our method is more successful in finding the correct salient pixels that SwAV. In many cases SwAV either fails to detect the more salient areas or focuses on the wrong pixels.

\begingroup
\setlength{\tabcolsep}{1pt} 
\renewcommand{\arraystretch}{1} 
\begin{table}[h]
    \centering
\begin{tabular}{c|c|c|c}
Algorithms & avg drop & avg increase  & dataset \\
\hline
SwAV+repo & 26.9  & 41.4 & Imagenet-1K \\
\hline
ours & \textbf{25.9}  & \textbf{45.2}  & Imagenet-1K\\
\hline
SwAV+repo & 30.2  & 36.6  & Imagenet-50 \\
\hline
ours & \textbf{29.2}  & \textbf{46.2} & Imagenet-50\\
\hline

\end{tabular}
\caption{Average Drop (less is better) and Average Increase (more is better). The models are trained on Imagenet-1K/Imagenet-50 $\texttt{train}$ and evaluated on Imagenet-1K/Imagenet-50 $\texttt{val}$ dataset.}
    \label{tab:quantitative}
\end{table}
\endgroup


\section{Conclusion}
In this paper we propose an add-on to the available self-supervised representation learning methods like \cite{caron2020unsupervised} by incorporating intra-class information including feature level location, and cross similarities between same class instances to the supervision signal. Experiments corroborate our theory and show effectiveness of our method in learning the additional local information included in the training signal.

\end{spacing}

{\small
\bibliographystyle{ieee_fullname}
\bibliography{references}
}

\clearpage

\clearpage
\newpage
\onecolumn
\setcounter{page}{1}
\setcounter{section}{0}

\section{Supplementary Materials}
This section contains supplementary materials that were initially omitted from the main body of the paper due to space limitations.

\subsection{More on the system model}
\paragraph{Clustering:} At the beginning of each epoc after computing the representation vectors we store them in the memory bank. This procedure is performed using the core engine. The core engine can be any self-supervised algorithm like \cite{caron2020unsupervised,he2019moco}.
\begin{figure}[htb!]
    \centering
    \includegraphics[width=10cm,height=5cm]{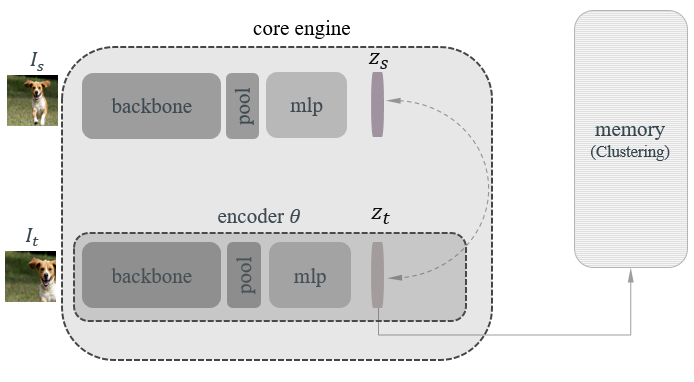}
    \caption{Schematics of the core representation learning engine and clustering memory. The representation vectors are stored in a memory bank at the beginning of each epoch.}
    \label{fig:core}
\end{figure}
\begin{figure}[htb!]
    \centering
    \includegraphics[width=12.5cm,height=5cm]{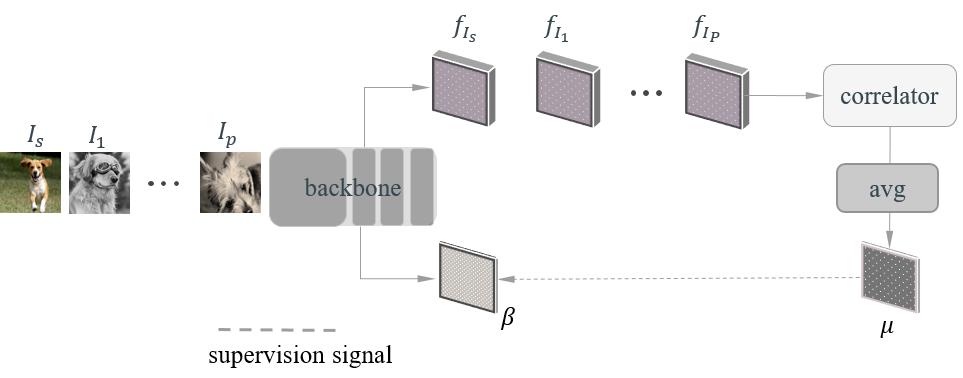}
    \caption{Attention mask $\beta$ predicts the correlation mask $\mu$ via a simple convolutional layer. The correlation between the feature maps $f$ of the images in the the set $\{I_s, \mathcal{P}\}$ generates a correlation mask $\mu$ that highlights key regions in the images from the same cluster. }
    \label{fig:attention}
\end{figure}
\paragraph{Positive sampling:}
After clustering the representations associated to the images in the datasets $\mathcal{I}$. We proceed to the next step which is positive sampling. For each image in the batch we sample $p$ same class co-images according to its clustering index.
\paragraph{Forming the attention-like masks:}
For each image in the class we would have $p$ positive samples associated to it. Therefore, a batch of size $N$ will contain $N\times(p+1)$ images.
We call the anchor image in the batch the source image $I_s$. We pass all the images in the batch from the backbone and get feature maps $f_{I_s}, f_{I_1},\ldots f_{I_p}$ associated to them. Then the attention supervision mask $\mu$ is created by averaging through the one-to-one correlation maps obtained by cross correlating the feature maps obtained in the previous step. We detach this supervision mask as we have to stop flowing the gradients back to the backbone by a supervision signal. This is critical because it can train a degenerative network.
\begin{figure}[htb!]
    \centering
    \includegraphics[width=12.5cm,height=5cm]{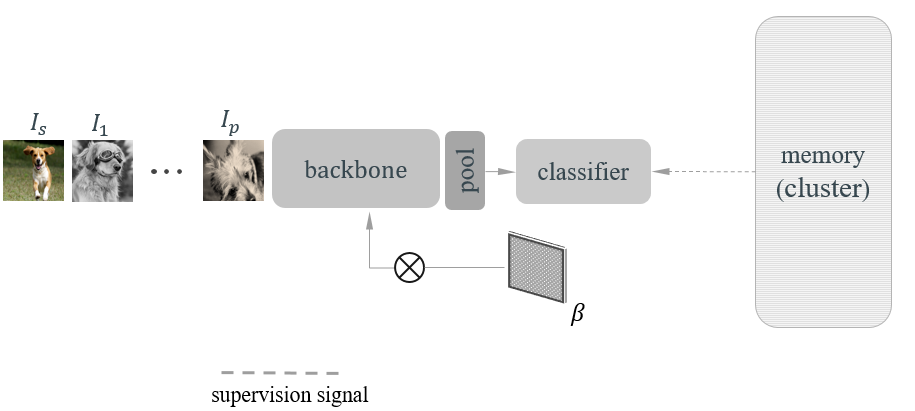}
    \caption{Attention mask $\beta$ is multiplied back to the network feature maps. The resultant feature maps are then passed through the rest of the network and a classification head to predict the class index they were originally picked from. }
    \label{fig:classification}
\end{figure}
A feature map $\beta$ is created on the other hand with the procedure we described in Sec.~\ref{sec_method}. We would like this signal to track the supervision attention-like mask $\mu$ in an MSE manner. Fig.~\ref{fig:attention} shows the details of this procedure.

In the next step we multiply $\beta$ to $f_I's$ and pass them through a classification head. We should be able to classify the explanation maps as the original classes they were picked from. ref. to Fig.~\ref{fig:classification}

Finally the network is trained by minimizing both the self-supervised loss and the attention mask MSE loss in \eqref{eq_final_loss}.

\subsection{Self-supervised-Pretraining}
In this section we describe our implementation details for Imagenet-1K \cite{deng2009imagenet} and imagenet-50 dataset. 
\paragraph{Datasets:}
We study the unsupervised learning in the realm of
Imagenet-1K \cite{deng2009imagenet} mainly. This training set contains more than 1.28 million images in 1000 classes. The image distribution over the classes in this dataset is uniform, and the subjects are fairly located in the center of each image.

We run our ablation studies on a randomly selected $50$-class subset of Imagenet-1K which we call it Imagenet-50. Since this distribution is sampled from Imagent-1K, the behaviour of our algorithm in this subset can fairly approximate its behaviour when exposed to Imagenet-1K. 
\paragraph{Core Engine: }
We use the SwAV \cite{caron2020unsupervised} structure as our core engine unless otherwise stated. As a common practice Resnet-50 (R-50) \cite{he2016deep}, is employed in our experiments.  
We replace the fully connected layer in Resnet-50 with a multi-layer-perceptron (MLP) architecture with two hidden layers to obtain the representation vectors.
Similar to the structure of SwAV the MLP projection head is a sequential concatenation of a fully connected layer, a batchnorm, and relu, followed by another fully connected layer \cite{caron2020unsupervised}. 
The output dimension of the MLP projection head is set to be $d=128$ as is in SwAV.

For a fair comparison to the SwAV base-line, we also use a linear head with weights of shape $2048\times 3000$, as our prototype kernel for Imagenet-1K training and $2048\times 150$, for Imagenet-50. The prototypes are responsible to map their corresponding representation vectors to maximal entropy codes of dimension $K=3000$ for Imagenet-1K and $K=150$ for Imagenet-50 \cite{caron2020unsupervised}.
\paragraph{Attention Module:}
The attention mask is created via a simple layer which is consist of a relu, followed by a linear fully connected layer with $\texttt{in\_channel} = C$, $\texttt{out\_channel}=1$, $\texttt{kernel\_size}=(1,1)$, and $\texttt{stride}=1$. During our experiments we found that if we append a Sigmoid layer to the architecture the convergence is smoother. The feature map $f$ is taken from the penultimate layer in the Res50 backbone which is of shape $N\times C \times 7\times7$, where $N$ is the batch size. We also experiment a multi-stage scenario where the feature map $f$ is a concatenation of multiple layers from the Res50 backbone. Each layer's feature map is interpolated to the size $28\times 28$ and then concatenated through the second axis, resulting in a feature map of size $N\times C \times 28 \times 28$, where $C$ is the aggregate channel counts.

The attention mask in our scenario tries to track the correlation mask in the MSE sense. Once the attention is predicted it is multiplied back to each channel of the feature map $f$ resulting in an attention imposed explanation map. The explanation map then continues to flow through the backbone and a pooling layer. From there the features are pushed through the classifier.
\paragraph{Classification Head:}
The classifier is consist of a fully connected layer with $K_c=3000$/$k_c=50$ neurons at the output for Imagenet-1K/Imagenet-50, respectively. This is equal to the k-means clustering number of classes.

\paragraph{Memory Bank:}
We register a memory bank for the whole dataset training data points. Each data point is associated with a representation vector of size $d=128$ which will consume $1.22 GB$ of GPU memory for a large dataset like Imagenet-1k. 
\paragraph{Hyper-parameters: }The temperature value $\tau_c$, is a hyper-parameter that plays an important role in our clustering scenario. Using higher values for $\tau_c$ results in softer probability distribution over the clusters. Since we deal with the cross-entropy loss across the cluster distributions generated by attention architectures, very small values of $\tau_c$ will diminish the importance of this loss, as the probabilities across clusters will be close to zero. Here we choose $\tau_c=.05$, unless otherwise stated. We also choose $K_c=3000$ in our Imagnet-1K experiments and $K_c=150$ for the ablation studies on Imagenet-50. Similarly, we set the SwAV number of prototypes $K=3000$ for Imagenet-1K and $K=150$ for Imagenet-50 studies. Other hyper-parameters that are related to the SwAV core engine, unless otherwise mentioned, are set to their default values as in \cite{caron2020unsupervised}.  We also set the loss weights to be $w_0=1,  w_1=.05,  w_2=.1$.
\paragraph{Unsupervised Training : }\label{sec_Unsupervised_training} 
We train the networks using LARC optimizer with momentum of .9 and weight decay of $1e-6$. We train on 64 Nvidia-V100 GPUs with mini-batch size of 2048 and a cosine learning rate schedule \cite{loshchilov2016sgdr} with coefficient of .5 and offset of .1. We use a 10-epoch warmup for both the core engine and the attention module. The starting learning rates, $lr_\texttt{core}=.3$ for the former and $lr_\texttt{attn}=.015$, for the later rise up to $lr_\texttt{core}=3.6$ and $lr_\texttt{attn}=.03$ at the 10th epoch and decay to  $lr_\texttt{core}=.0036$ and $lr_\texttt{attn}=.000001$, eventually. Training Resnet-50 for 200 epochs for the whole structure with the mentioned setting will take 296 hours.
\paragraph{Augmentation: }We use \texttt{RandomResizedCrop} in Pytorch, to crop and resize the the input image to two 224 x 224 crops. After that we use \texttt{RandomHorizontalFlip} with probability of .5. We also use a random color distortion composed of random \texttt{ColorJitter} with probability $80\%$ and (brightness=.8, contrast=.8, saturation=.8, hue=.2) and strength of $s=1$, followed by a \texttt{RandomGrayscale} with probability of $.2$, followed by a random Gaussian blur with $50\%$ probability and kernel size $23\times 23$ and uniform kernel with unit mean and variance of $.3$. We normalize the images at the last stage with mean = [0.485, 0.456, 0.406] and  std = [0.228, 0.224, 0.225].

The positive images also pass through a set of similar augmentations except that we generate only a single crop for each positive image sampled from the dataset.

\subsection{Classification, the Linear Head Protocol }\label{sec_linear_classification}
After the first stage of training, i.e.,  training the core-engine, the attention module, and the classifier in an unsupervised manner, we proceed to the second stage. In this part we take the backbone with a pooling layer attached to it, and replace the MLP structure with a fully connected layer of 1000-d/50-d, which is the number of classes in Imagenet-1K/Imagenet-50. We further freeze the weights of the backbone that were trained previously in the first stage, and only train the fully connected layer with the true Imagenet-1K/Imagenet-50 labels in a supervised fashion for 100 epochs. The philosophy behind the linear head training is that if in the first stage, the image semantics are learned properly, then in the second stage, the linear fully connected layer should learn to classify the images of the same dataset within a few epochs. We use Pytorch \texttt{CrossEntropyLoss} and \texttt{SGD} optimizer with momentum of $.9$ and weight decay of $.0001$ in this stage. We use batch size of 256 with initial learning rate of .3. We also use a cosine learning rate scheduler similar to Sec. \ref{sec_Unsupervised_training} for 100 epochs\cite{wang2015unsupervised}.

\subsection{Object Detection and Segmentation:}
We use Detectron2\cite{wu2019detectron2} framework to perform all object detection tasks by training Faster-RCNN\cite{ren2015faster} models. Specifically, we fine-tune all the parameters of the self-supervised network on PASCAL-VOC and COCO datasets. We report the VOC default metric of $\mathrm{AP_{50}}$, COCO style metric of $\mathrm{AP}$ and $\mathrm{AP_{75}}$, all averaged over $5$ trials. Unless otherwise stated all the experiments are run on $8$ NVIDIA-V100 GPUs with batch size of $16$.   

Due to limitations of the Batch Normalization (BN)\cite{ioffe2015batch} in structures like Resnet-50 it is relatively hard to achieve good detection results by training object detector from scratch \cite{he2019rethinking}\footnote{Object detectors are trained on images of higher resolution. This makes it inevitable to load batches of very small size to the GPUS due to memory limitations. Therefore, BN operates on very small batch sizes, resulting in inaccurate learned parameters\cite{ioffe2017batch,8578745,wu2018group}. }. Therefore, it is common to pre-train on Imagenet-1K in the supervised fashion, and then freeze the BN layers and fine-tune the detector on the downstream task thereby using the pre-trained network as the initialization. Detectron2 as a standard platform however, trains object detectors with selected standard hyper-parameters fine-tuned on Imagenet-1K supervised pre-trained networks. Since, detectors are highly dependant to their training hyper-parameters and the scheduling mechanism, to be fair in comparison, we follow the normalization routine adopted by \cite{he2019moco} via fine-tuning and synchronizing the BN trained across GPUs \cite{8578745}. This would bypass the need for freezing the BN layers trained on Imagenet-1K in the detector architecture \cite{he2016deep}. Also similar to \cite{he2019moco,chen2020exploring} we normalize the detector-specific appended layers by adding BN for better adjustment of the weights magnitude. As for fine-tuning, we use the same schedule as the Imagenet-1K supervised pre-training peer.    
\subsubsection{PASCAL-VOC Object Detection:}\label{sec_detect_full_voc}
We fine-tune all layers of a  Faster R-CNN \cite{ren2015faster} detector with a R50-C4 backbone on the VOC \texttt{trainval07+12} and evaluate on the VOC \texttt{test2007}. The image scale during training changes in the interval $[480~800]$ with the step of $32$ pixels, where at the test time is fixed on $800$ pixels. We train for $24k$ iterations where we reduce the learning rate by $\times10$ in $18k$ and $22k$ iteration marks. The initial learning rate is set as the default value of $.02$ with linear warm-up \cite{goyal2017accurate} for $100$ iterations. The weight decay and momentum are $0.0001$ and $0.9$, respectively. Table.~ \ref{tab:table_detection_pascal-R50-C4} shows the detection results fine-tuned on VOC dataset with R50-C4 backbone.

\begin{table}[bth!]
\centering
\begin{tabular}{c|c|c}
  & VOC 07 detection & VOC 07+12 detection\\

\hline
pre-train&  
\begin{tabular}{lll} $\textrm{AP}_{50}$ & $\textrm{AP}$ & $\textrm{AP}_{75}$ \end{tabular} &
\begin{tabular}{lll} $\textrm{AP}_{50}$ & $\textrm{AP}$ & $\textrm{AP}_{75}$ \end{tabular}\\
\hline

scratch&  
\begin{tabular}{lll} 35.9 & 16.8 & 13.0 \end{tabular}&
\begin{tabular}{lll} 60.2 & 33.8 & 33.1 \end{tabular}\\
\hline

Imagenet-1K-pretraining&  
\begin{tabular}{lll} 74.4& 42.4& 42.7 \end{tabular}&
\begin{tabular}{lll} 81.3& 53.5& 58.8 \end{tabular}\\
\hline

SwAV+repr&  
\begin{tabular}{lll} 72.9 & 39.2 & 37.4 \end{tabular}&
\begin{tabular}{lll} 74.9 & 42.6 & 42.4 \end{tabular}\\
\hline

ours&  
\begin{tabular}{lll} \textbf{73.2} & \textbf{40.1} & \textbf{38.9} \end{tabular}&
\begin{tabular}{lll} \textbf{75.3} & \textbf{43.1} & \textbf{43.4} \end{tabular}\\
\hline
\end{tabular}
\caption{Transfer Learning. VOC 07 detection: Faster
R-CNN fine-tuned in VOC 2007 trainval, evaluated in VOC 2007 test; VOC 07+12 detection: Faster R-CNN fine-tuned in VOC 2007
trainval + 2012 train, evaluated in VOC 2007 test}
\label{tab:table_detection_pascal-R50-C4}
\end{table}

\subsubsection{COCO Object Detection and Segmentation:}\label{sec_detect_full_COCO}
Similar to \cite{chen2020exploring} we use a Mask-RCNN \cite{he2017mask} with C4 backbone. We train for $180k$ iterations with stepping down the learning rate by $\times 10$ at the iterations $120k$ and $160k$ (The so called $\times 2$ schedule routine). We fine tune all layers on \texttt{train2017} and validate on \texttt{val2017} sets. The image size for training is within [640, 800] and at the inference is 600. The batch size is 16 and the learning rate is the default value of $.02$.

\begin{table}[bth!]
\centering
\begin{tabular}{c|c|c}&COCO detection & COCO instance seg.\\

\hline
pre-train&  
\begin{tabular}{ccc} $\textrm{AP}_{50}$ & $\textrm{AP}$ & $\textrm{AP}_{75}$ \end{tabular}&
\begin{tabular}{ccc} $\textrm{AP}_{50}^{\textrm{mask}}$ & $\textrm{AP}^{\textrm{mask}}$ & $\textrm{AP}_{75}^{\textrm{mask}}$ \end{tabular}\\
\hline

scratch&  
\begin{tabular}{ccc} 44.0 & 26.4 & 27.8 \end{tabular}&
\begin{tabular}{ccc} $~~~46.9~~~$ & $~~~29.3~~~$ & $~~~30.8~~~$ \end{tabular}\\
\hline

Imagenet-1K-pretraining&  
\begin{tabular}{ccc} 58.2& 38.2& 41.2 \end{tabular}&
\begin{tabular}{ccc} $~~~54.7~~~$ & $~~~33.3~~~$ & $~~~35.2~~~$ \end{tabular}\\
\hline

SwAV+repr&  
\begin{tabular}{ccc} 59.4 & 39.1 & 42.3 \end{tabular}&
\begin{tabular}{ccc} $~~~55.6~~~$ & $~~~34.1~~~$ & $~~~36.0~~~$ \end{tabular}\\
\hline

ours&  
\begin{tabular}{ccc} \textbf{59.6} & \textbf{39.5} & 42.3 \end{tabular}&
\begin{tabular}{ccc} $~~~\textbf{55.9}~~~$ & $~~~\textbf{35.0}~~~$ & $~~~36.0~~~$ \end{tabular}\\
\hline
\end{tabular}
\caption{{Transfer Learning. All unsupervised methods are based on 200-epoch pre-training in Imagenet-1K. COCO detection and COCO instance segmentation: Mask R-CNN (2$x$ schedule) with C4-backbone,
fine-tuned in COCO 2017 train, evaluated in COCO 2017 val.}}
\label{tab:coco_detection}
\end{table}

\begin{figure*}[htb!]
\centering
\begin{subfigure}[a]{.6\textwidth}
   \includegraphics[width=1\linewidth]{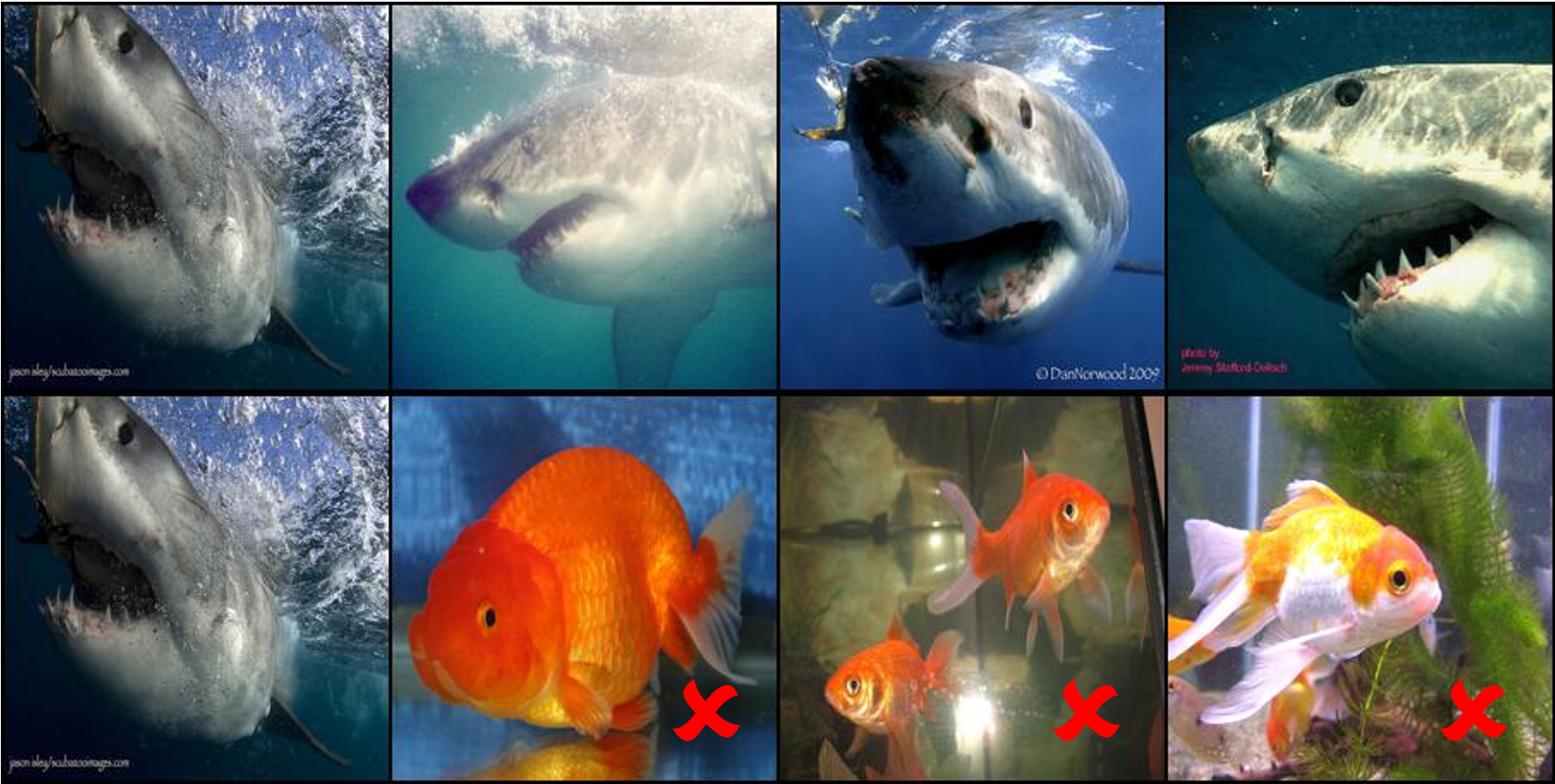}
   \caption{}
   \label{fig:knn_query_1} 
\end{subfigure} \qquad \qquad
\begin{subfigure}[b]{.6\textwidth}
   \includegraphics[width=1\linewidth]{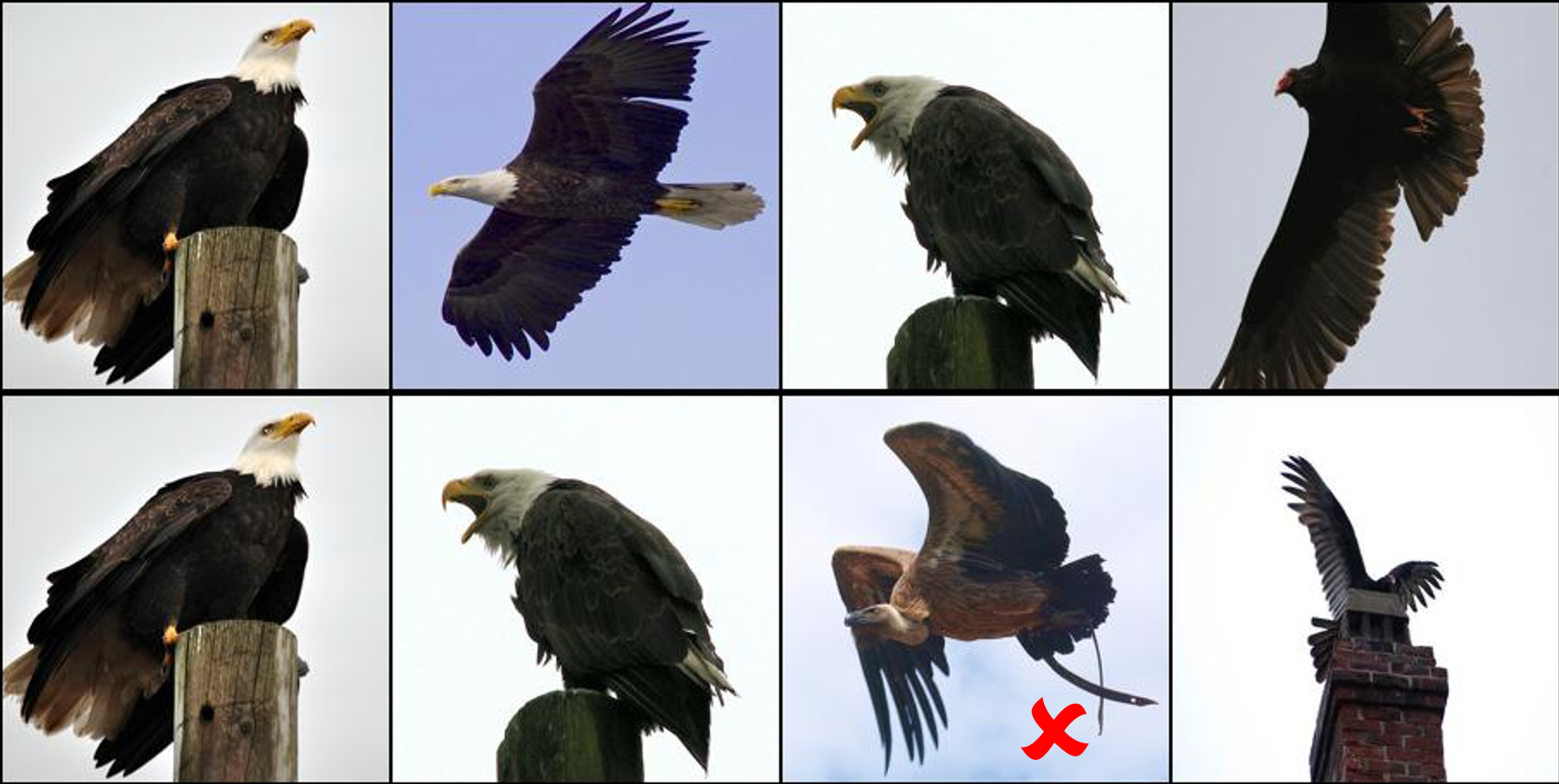}
   \caption{}
   \label{fig:knn_query_2}
\end{subfigure}
\begin{subfigure}[c]{.6\textwidth}
   \includegraphics[width=1\linewidth]{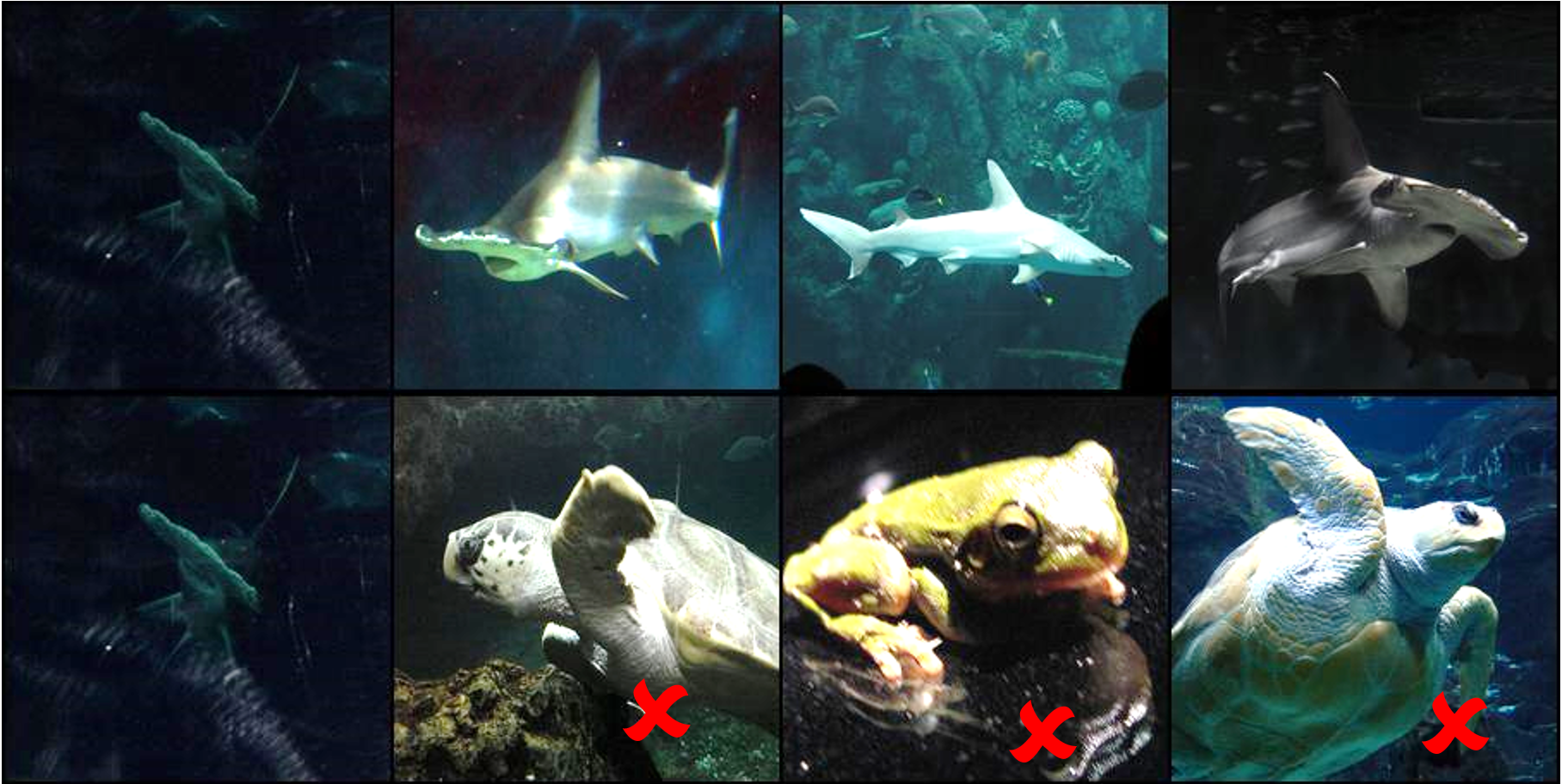}
   \caption{}
   \label{fig:knn_query_3}
\end{subfigure}
\caption{Results of positive query, anchor images (left column): Top (ours) vs bottom (SwAV). This figure shows the top-3 most similar set to the anchor images in the validation set, in terms of cosine similarity.}
\label{fig:knn_query}
\end{figure*}

\subsection{K-NN Image Query}
In this section we compare the performance of our algorithm with respect to the baseline SwAV qualitatively. For this, after pre-training on Imagenet-1K we detach the mlp layers and get our features from the Resnet-50 pooling layer. Applying K-NN to the validation set we cluster our validation set into overlapping classes. We divide this experiment into two categories:
\begin{enumerate}
    \item \textbf{Positive Query:}In this context we look for the top-k (k=3 here) nearest neighbors for some random anchor image. Fig.~\ref{fig:knn_query} shows the results of SwAV vs our algorithm. The images in the left column are the anchor images. The images in each row are top-3 nearest neighbors in the validation dataset. As it can be seen our algorithm outperforms SwAV in picking the most similar images from the validation set.
    \item \textbf{Negative Query:} In this experiment we pick the top-20 nearest neighbors to the anchor images to form a query set. Then we pick 3 least similar images in terms of cosine similarity to each anchor from the query set. The better algorithm is the one with most similar images to the anchor. 
\end{enumerate}

\begin{figure*}[htb!]
\centering
\begin{subfigure}[a]{.6\textwidth}
   \includegraphics[width=1\linewidth]{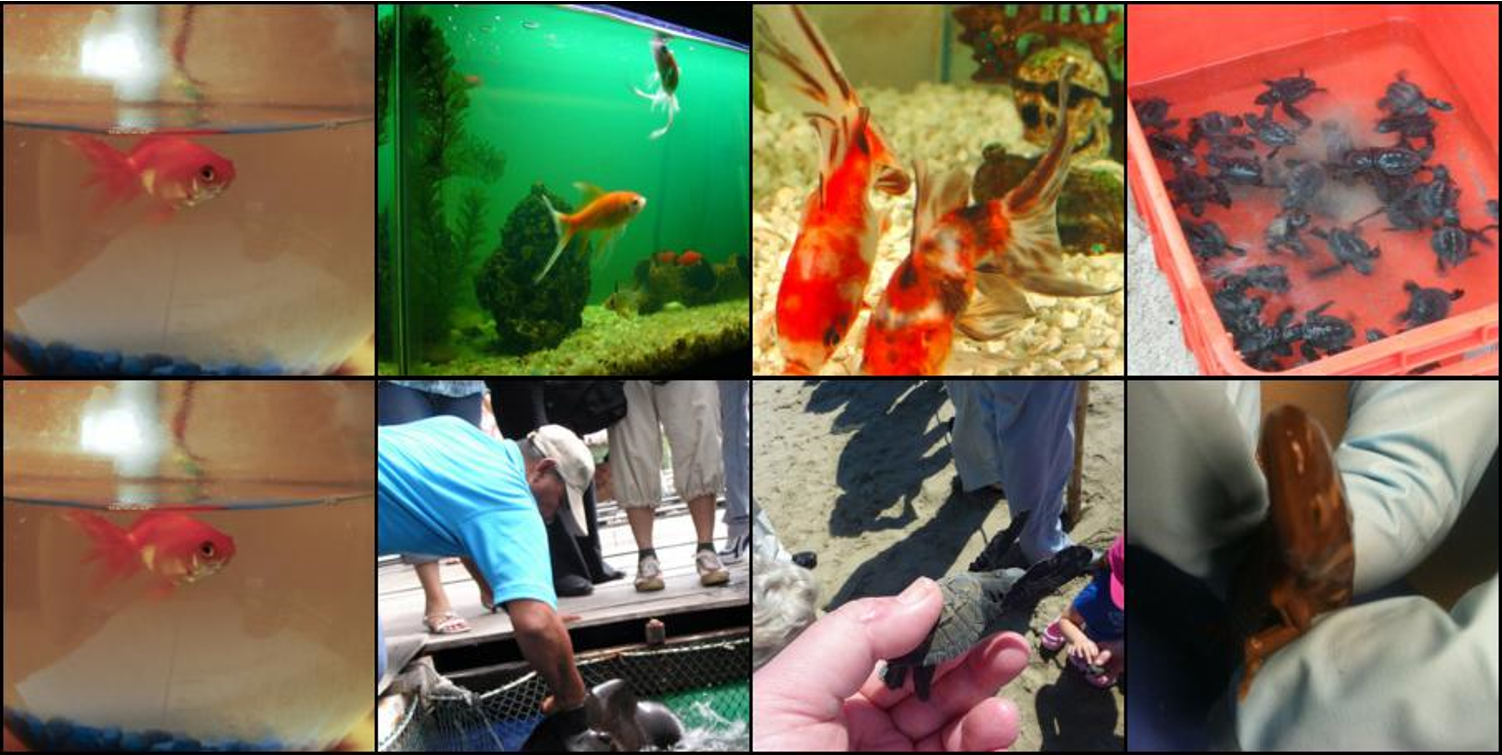}
   \caption{}
   \label{fig:nknn_query_1} 
\end{subfigure} \qquad \qquad
\begin{subfigure}[b]{.6\textwidth}
   \includegraphics[width=1\linewidth]{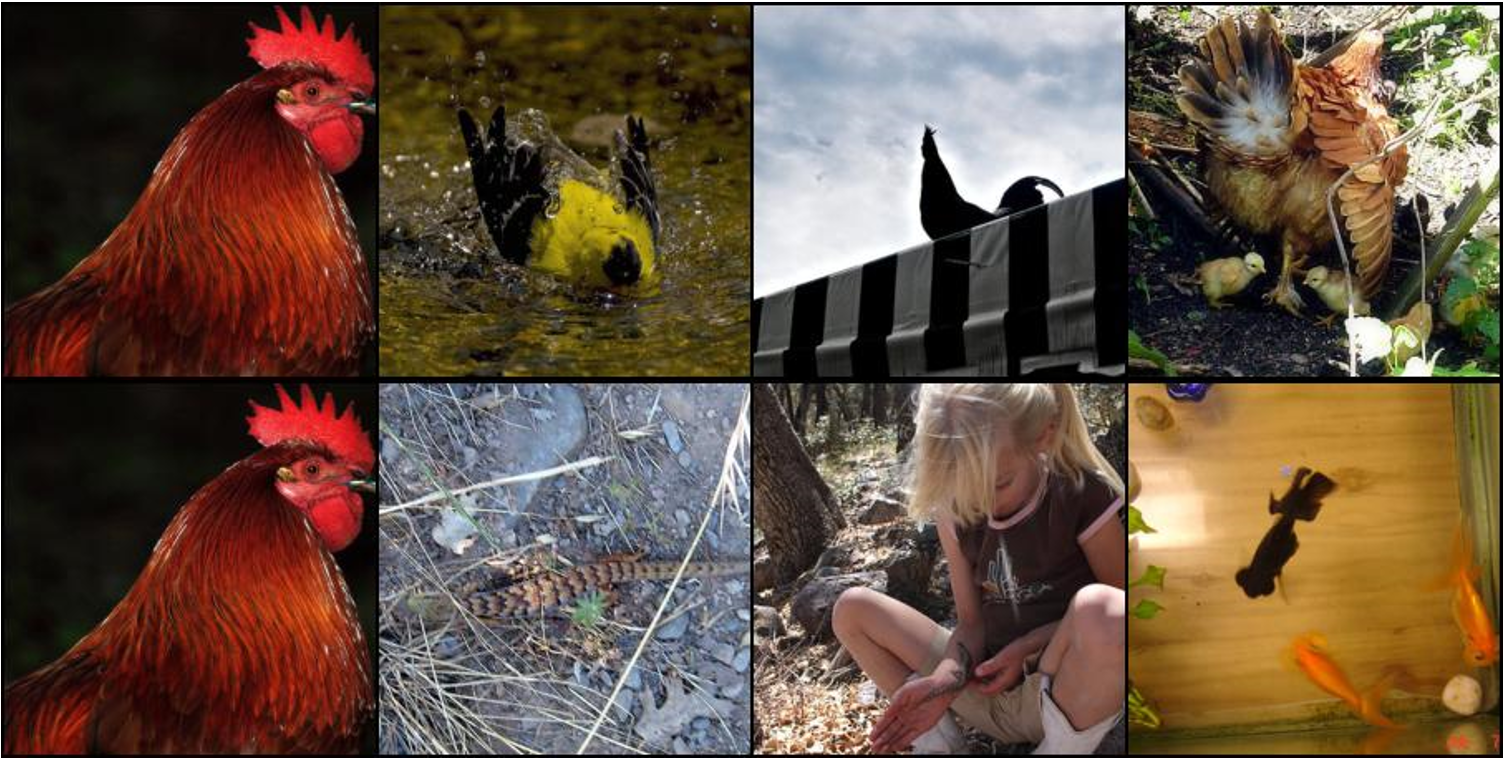}
   \caption{}
   \label{fig:nknn_query_2}
\end{subfigure}
\begin{subfigure}[c]{.6\textwidth}
   \includegraphics[width=1\linewidth]{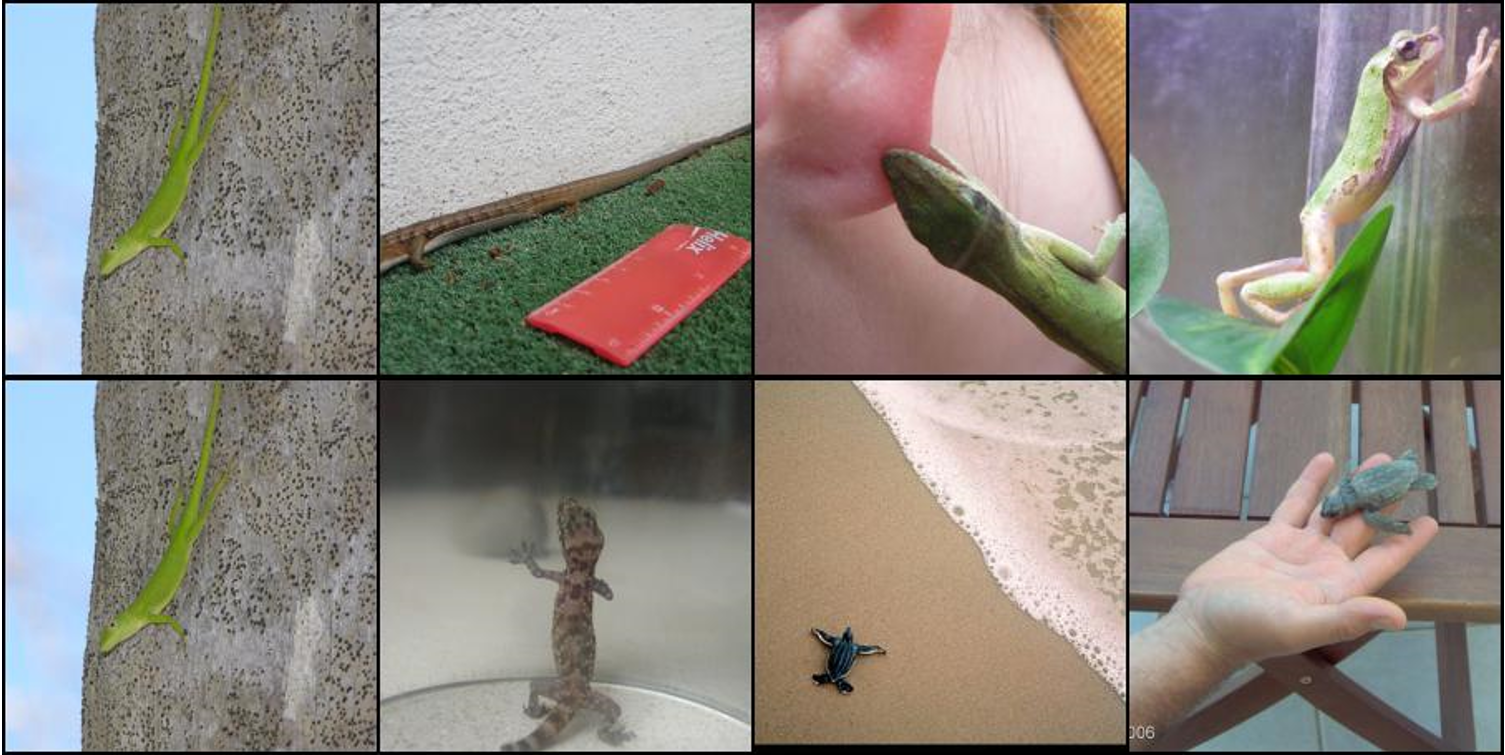}
   \caption{}
   \label{fig:nknn_query_3}
\end{subfigure}
\caption{Results of negative querying, anchor images (left column): Top (ours) vs bottom (SwAV). This figure shows the top-3 least similar images clustered in the same class by the K-NN algorithm for K=20.}
\label{fig:nknn_query}
\end{figure*}


\end{document}